%% file: root.tex
\documentclass[runningheads]{llncs}
\usepackage{graphicx}
\usepackage{comment}
\usepackage{amsmath,amssymb} 
\usepackage{color}

\usepackage{adjustbox}
\usepackage{multirow, booktabs} 
\usepackage{lipsum}

\newcommand{\secref}[1]{Sec.~\ref{#1}}
\newcommand{\figref}[1]{Fig.~\ref{#1}}

\newcommand{\tableref}[1]{Table~\ref{#1}}

\begin{document}
\pagestyle{headings}
\mainmatter
\def\ECCVSubNumber{xxxx}  

\title{PiP: Planning-informed Trajectory Prediction for Autonomous Driving}

\titlerunning{PiP: Planning-informed Trajectory Prediction}
%
\author{Haoran Song\inst{1} \and
Wenchao Ding\inst{1} \and
Yuxuan Chen\inst{2}  \and
Shaojie Shen\inst{1} \and \\
Michael Yu Wang\inst{1} \and
Qifeng Chen\inst{1}
}
\authorrunning{H. Song et al.}
%
\institute{The Hong Kong University of Science and Technology \\
\and
University of Science and Technology of China\\
}

\maketitle

\input{0_abstract.tex}
\input{1_introduction.tex}
\input{2_relatedwork.tex}
\input{3_method.tex}
\input{4_experiments.tex}
\input{5_conclusion.tex}

%
%
\clearpage
\bibliographystyle{splncs04}
\bibliography{ref}
\end{document}

%% file: 0_abstract.tex

\begin{abstract}
It is critical to predict the motion of surrounding vehicles for self-driving planning, especially in a socially compliant and flexible way.
However, future prediction is challenging due to the interaction and uncertainty in driving behaviors.
We propose planning-informed trajectory prediction (PiP) to tackle the prediction problem in the multi-agent setting.
Our approach is differentiated from the traditional manner of prediction, which is only based on historical information and decoupled with planning.
By informing the prediction process with the planning of ego vehicle, our method achieves the state-of-the-art performance of multi-agent forecasting on highway datasets.
Moreover, our approach enables a novel pipeline which couples the prediction and planning, by conditioning PiP on multiple candidate trajectories of the ego vehicle, which is highly beneficial for autonomous driving in interactive scenarios.


\keywords{Trajectory prediction $\cdot$ Autonomous driving}

\end{abstract}

%% file: 1_introduction.tex
\section{Introduction}

Anticipating future trajectories of traffic participants is an essential capability of autonomous vehicles. Since traffic participants (agents) will affect the behavior of each other, especially in highly interactive driving scenarios, the prediction model is required to anticipate the \textit{social interaction} among agents in the scene to achieve socially compliant and accurate prediction.


Despite the fact that the interaction among traffic agents is being investigated, far less attention is paid to how the uncontrollable (surrounding) agents interact with the controlled (ego) agent.
Different future plans of the ego agent will largely affect the future behaviors of all surrounding agents, which leads to a significant difference in future predictions.
Human drivers are accustomed to imagining \textit{what} the situation will be \textit{if} they are going to act in different ways.
For example, they speculate whether the other vehicles will leave space if they insert aggressively or mildly, respectively.
By considering the different future situations from multiple ``\textit{what-ifs}'', human drivers are adept at negotiating with other traffic participants while flexibly adapting their own driving behaviors.
The key is that human drivers condition the prediction of surrounding vehicles on their own future intention.
In this paper, we want to inform the interaction-aware prediction using the candidate plans of the controlled vehicle to mimic this thinking process.

\begin{figure}[t]
\begin{center}
\includegraphics[width=0.99 \linewidth]{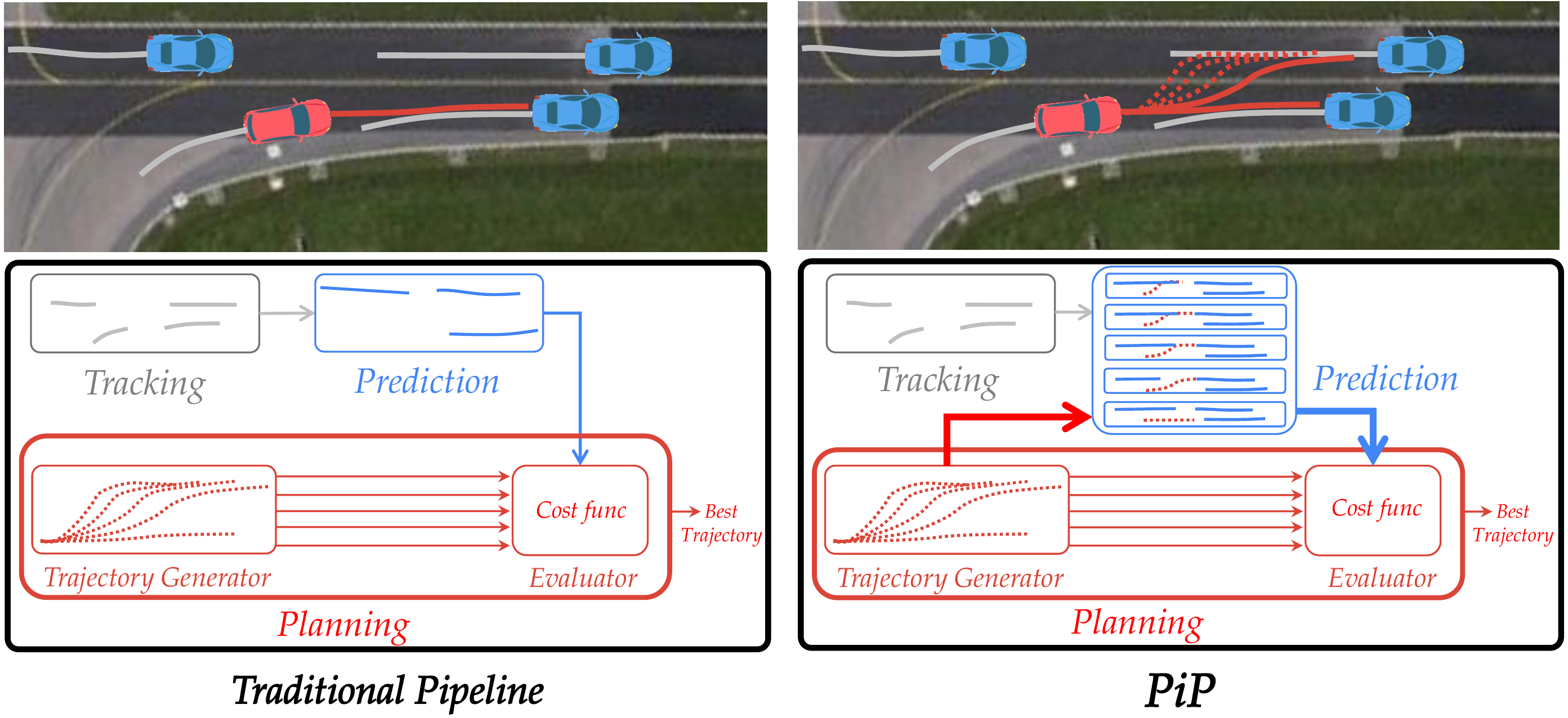}
\end{center}
\vspace{-0.1cm}
  \caption{Comparison between the traditional prediction approach (left) and PiP (right) under a lane merging scenario.
Assume the ego vehicle (red) intends to merge to the left lane. It is required to predict the trajectories of surrounding vehicles (blue).
To alleviate the uncertainty led by future interaction, PiP incorporates the future plans (dotted red curve) of ego vehicle in addition to the history tracks (grey curve).
While the traditional prediction result is produced independently with the ego's future, PiP produces predictions one-to-one corresponding to the candidate future trajectories by enabling the novel planning-prediction-coupled pipeline.
Therefore, PiP evaluates the planning safety more precisely and achieves more flexible driving behavior (solid red curve) compared with the traditional pipeline.
  }
\label{fig:overview}
\vspace{-0.1cm}
\end{figure}

To this end, we propose a novel planning-informed prediction framework (PiP).
Note that PiP does not require the exact future trajectory, which is actually undetermined during prediction.
PiP only conditions the prediction process on the candidate future trajectories proposed by the trajectory generator, like ``insert aggressively'' and ``insert mildly'' these kinds of ``what-ifs''.
Accordingly, the best trajectory could be picked out after evaluating all the candidate plans by their corresponding predictions in the planning module.

There are two significant benefits of PiP.
First, by incorporating the additional planning information, the interaction among agents can be better captured, which leads to a considerable improvement in predictive accuracy.
Second, the planning-informed prediction will provide a highly valuable interface for the planning module during system integration.
Explicitly, instead of evaluating multiple future plans under a fixed prediction result as most autonomous driving systems do,
PiP conditions the prediction process on the ego vehicle's future plans, which uncovers how the other vehicles will interact with ego vehicle if the ego vehicle executes any specific planning trajectory.
The PiP pipeline is especially suitable for planning in dense and highly interactive traffic (such as merging into a congested lane), which is hard to be handled using traditional decoupled prediction and planning pipeline.
The comparison between the traditional pipeline for autonomous driving and PiP is illustrated in \figref{fig:overview}.

To effectively achieve planning-informed prediction, we propose two modules, namely, planning-coupled module and target fusion module. The planning-coupled module extracts the interaction features with a special channel for injecting the future planning, while the target fusion module encodes and decodes the tightly coupled future interaction among agents. PiP is end-to-end trainable.
The main contributions are listed as follows:
\begin{itemize}
\renewcommand{\labelitemi}{\textbullet}
    \item \textit{Planning-coupled} module is proposed to model the multi-agent interaction from both the history time domain (history tracking of surrounding agents) and future time domain (future planning of controlled agent). By introducing the planning information into social context encoding, the uncertainty from the multi-modality of driving behavior is alleviated and thus leads to an improvement in predictive accuracy.
    \item \textit{Target fusion} module is presented to capture the interdependency between target agents further. Since all the future states of targets are linked up together with the planning of the controlled agent, we apply a fully convolutional structure to model their future dependency at different spatial resolutions. The introduction of the target fusion module leads to further improvement for multi-agent forecasting.
    \item Our model outperforms state-of-the-art methods for multi-agent forecasting from tracking data.
    Moreover, the proposed planning-prediction-coupled pipeline extends the operational domain of planning by the integration with prediction, and some qualitative results are demonstrated.
\end{itemize}

%% file: 2_relatedwork.tex

\section{Related Work} \label{sec:related_work}
To accurately forecast the future trajectory of a specific vehicle, we need to discover the clues from its past observation and corresponding traffic configuration. In this paper, we focus on the data-driven trajectory prediction methods, which essentially learn the relationship between future trajectory and past motion states. Since vehicle behaviors are often inter-related, especially in dense traffic, it is crucial to consider interaction-aware trajectory prediction for autonomous driving, namely, in a multi-agent setting. In this section, we provide an overview of interaction-aware trajectory prediction methods and the common practice of integrating prediction with planning, which motivates our planning-informed prediction.

\textbf{Interaction-aware trajectory prediction:} Multi-agent learning and forecasting~\cite{felsen2018will,le2017coordinated,lee2016predicting,sun2019stochastic,zhan2018generative} is a challenging problem and Social LSTM~\cite{alahi2016slstm} is one seminal work. In~\cite{alahi2016slstm}, the spatial interaction among pedestrians is learned using the proposed social pooling structure based on the hidden states generated by long short-term memory (LSTM) network, and~\cite{deo2018cslstm} improves the social pooling strategy by applying convolutional layers. To better capture the multi-modal nature of future behaviors, some non-deterministic generative models are adopted based on generative adversarial network (GAN)~\cite{goodfellow2014gan,gupta2018sgan,sadeghian2019sophie}, and variational autoencoder (VAE) \cite{kingma2013auto,lee2017desire}. Besides learning the interaction among agents, the agent-scene interaction is also modeled in~\cite{bartoli2018context,sadeghian2018car,zhao2019matf}. The interaction-aware network structures are further extended to heterogeneous traffic~\cite{chandra2019traphic,ma2019trafficpredict} and applied to autonomous driving scenarios such as~\cite{deo2018cslstm,ding2019extended,lee2017desire}.


\textbf{Trajectory prediction for control and planning:}  
Targeting on the real-time driving, the popularly used vehicle motion planners \cite{fan2018baidu,mcnaughton2011motion,pivtoraiko2009differentially,schwarting2018planning,werling2010optimal} follow the workflow: 
first roll out multiple candidate ego trajectories; 
then score them using user-defined functions, in which the future trajectories of other vehicles predicted based on history tracks are considered; 
finally, pick out the best trajectory to execute. 
Note that the prediction result of other vehicles is fixed for different candidates from the trajectory generator of the ego vehicle. 
Namely, the traditional pipeline does not make ``what-ifs'', and think the reactions of other vehicles will be the same even given different ego actions.
However, because the future planning of the ego vehicle in turn affects the behaviors of surrounding agents, the ``predict-and-plan'' workflow may be inadequate, especially in tightly coupled driving scenarios such as merging~\cite{hubmann2018belief}. Differentiated from the traditional decoupled pipeline, PiP can be incorporated into a novel planning-prediction-coupled pipeline, which extends flexibility in dense traffic. 

\textbf{Planning-informed trajectory prediction:}
Rhinehart~\cite{rhinehart2019precog} proposed conditioning the prediction process on the goal of the ego vehicle, which is likely to be the first attempt to incorporating planning information into prediction. However, only the goal position information is utilized, which may pose restrictions. 
For example, considering the scenario of controlling the ego vehicle to insert into congested traffic, even given the same goal of the ego vehicle, the future reaction of surrounding vehicles varies significantly depending on the specific time profile of how the ego vehicle reaches the goal. 
This motivates us to inform the prediction process with varying completeness of planning information (i.e., from several rough waypoints to a complete planned trajectory). 
In general, our proposed method is capable of providing accurate interaction-aware trajectory prediction for a large batch of different candidate planned trajectories efficiently, which facilitates planning in highly interactive environments~\cite{cunningham2015mpdm,ding2019safe}.

%% file: 3_method.tex

\section{Method}
\label{sec:method}
In PiP, the motion of each target vehicle is predicted by considering not only its own state and the other agents' states in the history time domain, but also the ego vehicle's planned trajectory.
In this section, we first formulate the problem in~\secref{subsec:formulation}, and describe the details of PiP in the following structure:
the planning-coupled module which incorporates the ego vehicle's planned trajectory in the social tensors of neighboring vehicles' past motions (\secref{subsec:planning_coupled}), the method of agent-centric target fusion (\secref{subsec:target_fusion}) and the maneuver-based decoding method for generating the probabilistic distribution of the location displacement between future frames (\secref{subsec:maneuver_decoding}). Some implementation details are provided in~\secref{subsec:implementation}.

\subsection{Problem Formulation}
\label{subsec:formulation}
Consider the driving scenario for an autonomous vehicle. The ego vehicle is commanded by the planning module, and the perception module senses the neighboring vehicles within a certain range.
We formulate the trajectory prediction problem in the multi-agent setting as estimating the future states of a set of target vehicles around the ego vehicle $v_{ego}$ conditioning on the tracking history of all surrounding vehicles and the planned future of the controllable ego vehicle.
The objective is to learn the posterior distribution $P(\mathbf{Y}|\mathbf{X}, \mathcal{I})$ of multiple targets' future trajectories $\mathbf{Y} = \left\{Y_i | v_i \in {V}_{tar} \right\}$, where ${V}_{tar}$ is the set of predicted targets selected within an ego-vehicle-centric area ${A}_{tar}$.
The conditional items contain the future planning of ego vehicle $\mathcal{I}$ and the past trajectories $\mathbf{X} = \left\{X_i | v_i \in V \right\}$, where $V$ denotes the set of all vehicles involved around the ego vehicle,
and $(v_{ego} \cup V_{tar}) \subseteq V$ as the ego vehicle is not required to be predicted.
At any time $t$, the history trajectory and future trajectory of an agent $i$ are denoted as $X_i=\left\{ x_i^{t-T_{obs}+1}, x_i^{t-T_{obs}+2} ..., x_i^t \right\}$
and $Y_i=\left\{ y_i^{t+1}, y_i^{t+2} ..., y_i^{t+T_{pred}} \right\}$, where the elements of $x_i, y_i \in \mathbb{R}^2$ represent waypoint coordinates in the past and future, respectively, while $T_{obs}$ and $T_{pred}$ refer to the number of frames for observation and prediction.
Note that the planned trajectory $\mathcal{I} = Y_{ego} = \left\{ y_{ego}^{t+1}, y_{ego}^{t+2} ..., y_{ego}^{t+T_{pred}} \right\}$is also used as a conditional item, since it's generated from ego vehicle's trajectory planner and thus can be accessible during prediction.
Moreover, the introduction of $\mathcal{I}$ enables the planning-prediction-coupled pipeline as shown in Fig.~\ref{fig:overview}.

\begin{figure*}[t]
\begin{center}
\includegraphics[width=1.0\linewidth]{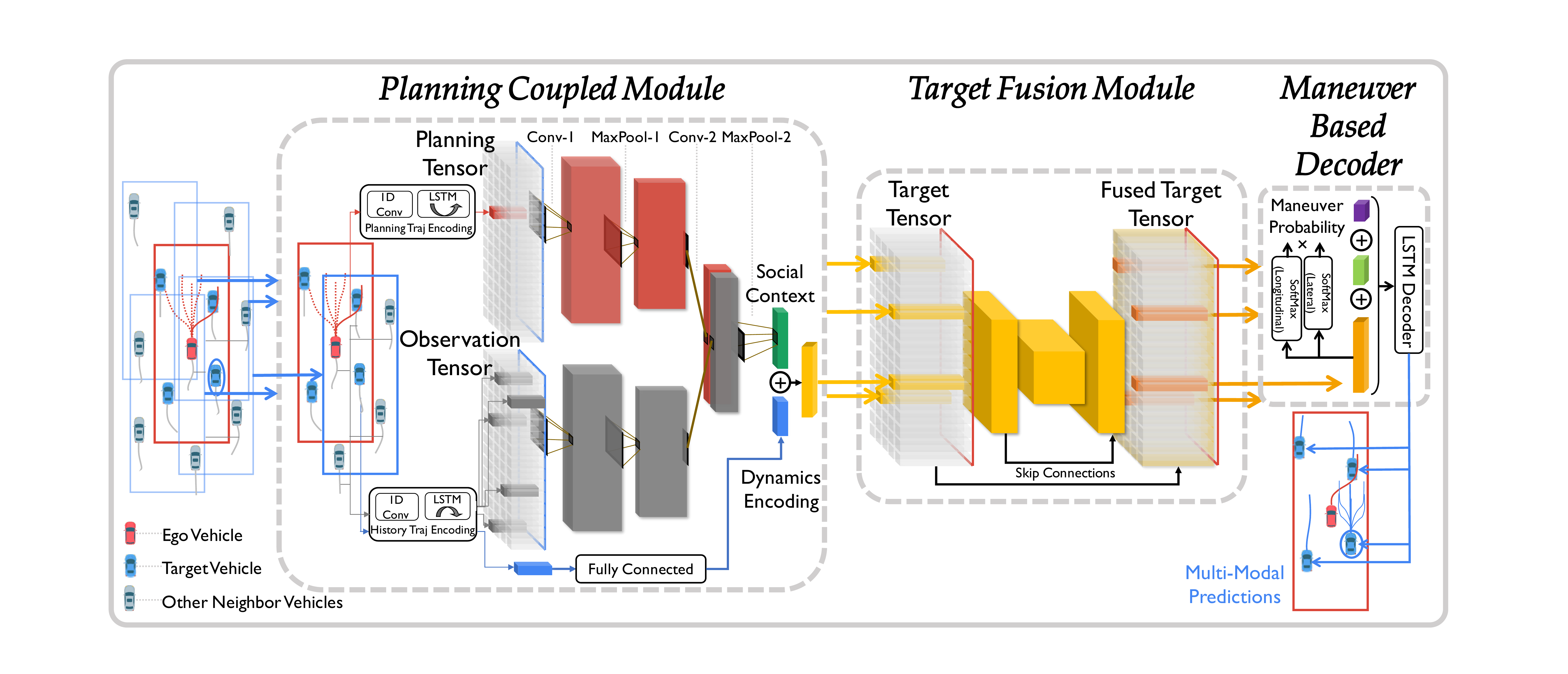}
\end{center}
  \caption{The overview of \textit{PiP} architecture: \textit{PiP} consists of three key modules, including planning coupled module, target fusion module, and maneuver-based decoding module. Each predicted target is firstly encoded in the planning coupled module by aggregating all information within the target-centric area (blue square). A target tensor is then set up within the ego-vehicle-centric area (red square) by placing the target encodings into the spatial gird based on their locations. Afterward, the target tensors are passed through the following target fusion module to learn the interdependency between targets, and eventually, a fused target tensor is generated. Finally, the prediction of each target is decoded from the corresponding fused target encoding in the maneuver-based decoding module. The target vehicle marked with an ellipse is exemplified for planning coupled encoding and multi-modal trajectories decoding.
  }
\label{fig:architecture}
\end{figure*}



\subsection{Planning Coupled Module}
\label{subsec:planning_coupled}
In the planning coupled module, each predicted agent is processed in its own centric area $\mathcal{A}_{nbr}$, in which the ego vehicle $v_{ego}$, the target vehicle $v_{cent} \in {V}_{tar}$ and the other neighboring vehicles $V_{nbrs} \subseteq {V}$ located within $\mathcal{A}_{nbr}$ are included.
There involve three encoding streams: the dynamic property of the target itself, the social interaction with the target's neighboring vehicles, and the spatial dependency with ego vehicle's future planning. Consequently, a target encoding $\mathcal{T}$ is generated by embedding these encodings together.
In practice, we use relative trajectories in an agent-centric manner for capturing interdependencies between the centric agent and surrounding agents.

\textbf{Trajectory Encoding:}
All trajectories contained in the planning coupled module could be classified into two types: observable and controllable.
The history trajectories of traffic participants could be observed, and the planned trajectory to command the ego vehicle could be controlled.
Before extracting the spatially interactive relationship between traffic agents, all trajectories are encoded independently to learn the temporal properties in their sequential locations.
To better accomplish this work, each trajectory is preprocessed by converting its locations into relative coordinates with respect to the target vehicle and then fed into a temporal convolutional layer to obtain a motion embedding. After that, the Long Short-Term Memory (LSTM) networks are employed to encode the motion property for trajectories, and the hidden state $h(\cdot)$ therein is regarded as the motion encoding for the corresponding trajectory. Here, the LSTMs with different parameters are adopted for planned trajectory $Y_{ego}$ and history trajectories including $X_{ego}$, $X_{cent}$ and $X_{nbr}$, as they belong to the different time domains.

\textbf{Planning and Observation Fusion:}
The use of LSTM encoder captures the temporal structure from the trajectory sequence, while it fails to handle the spatial interaction relationship with other agents in a scene.
The social pooling strategy, proposed in~\cite{alahi2016slstm}, addresses this issue by pooling LSTM states of spatially proximal sequences in a target-centric grid named as ``social tensor''.
The ``convolutional social pooling'' in~\cite{deo2018cslstm} improves the strategy further by applying convolutional and max-pooling layers over the social tensor. Both of the methods are proposed for learning the spatial relationship among trajectories that takes place in the history period.
In our proposed framework, we adopt the convolutional social pooling structure for modeling spatial interaction.
In addition to interdependencies between target and neighbors in the past time, the spatial information of ego vehicle's planning in the future time is counted in the planning coupled module as an improvement.
Accordingly, three encoding branches stemming from LSTM hidden states of all trajectories are included, as illustrated in \figref{fig:architecture}.
The lower branch encodes the dynamics property of the target vehicle by feeding its motion encoding $h(X_{cent})$ to a fully connected layer.
The spatial relationship between the target and its surrounding agents is captured in the upper branches by building a grid centered at the location of the target vehicle.
Since the planned future trajectory and observed history trajectory belong to different time domain,  the history information of $h(X_{nbr})$ and $h(X_{ego})$ are placed into a target-centric spatial grid termed as observation tensor with respect to the corresponding locations at current time $t$, while the motion encoding of the planned trajectory $h(Y_{ego})$ is placed similarly in another spatial grid to form the planning tensor. It should be noted that the planning sequence is encoded in a reversed order because the planning of the near future is more reliable, and thus it should weight more in the encoding.

After that, both of the observation and planning tensors pass through convolutional layers and pooling layers in parallel and then are concatenated together before fed to the last max-pooling layer.
Merging the information from the planning of ego vehicle and observation of surrounding vehicles, the resulting encoding $\mathcal{S}$ covers the social context for both the past and future time domain.
Finally, the merged social encoding $\mathcal{S}$ concatenates with the target's dynamics encoding $\mathcal{D}$ to form a target encoding $\mathcal{T}$ that aggregates all the information accessible within the target-centric grid.

\subsection{Target Fusion Module}
\label{subsec:target_fusion}
In~\cite{alahi2016slstm,deo2018cslstm}, the future states of each target is directly decoded from the agent-centric encoding result that aggregates history information.
In this way, each trajectory is generated independently from the corresponding target encoding.
However, the future states of targets are highly correlated, which indicates that the decoding process for a certain target also depends on the encoding of other targets. Therefore, we further fuse the encoding among different targets in the scene and decode the final trajectory from the fused encoding, which better captures the dependencies of future states of different targets in the same scene.

For jointly predicting the vehicles within the target area centered on the ego vehicle's location, each target vehicle $v_i \in {V}_{tar}$ represented by its encoding $\mathcal{T}_i$ is placed into an ego-vehicle-centric grid $\left\{ \mathcal{T}_i | v_i \in V_{tar} \right\}$ based on their locations at the last time step of history trajectories.
Inspired by some popular CNN architectures for segmentation~\cite{long2015fcn,ronneberger2015} that produce correspondingly-sized output with hierarchical inference, we adopt the fully convolutional network (FCN) to learn the context of target tensor.
The target tensor is fed into a symmetric FCN structure for capturing the spatial dependencies between target agents at different grid resolutions, where the skip-connected layers are combined by element-wise sum.
The fused target tensor produced by this module retains its spatial structure the same as before fusion, from which the fused target encoding $\mathcal{T}^{+}_i$ of each target could be sliced out according to its grid location.

\subsection{Maneuver Based Decoding}
\label{subsec:maneuver_decoding}
To address the inherent multi-modality nature of the driving behaviors, the maneuver based decoder built upon~\cite{deo2018cslstm} is applied to predict the future trajectory for predefined maneuver classes $M=\left \{ m_k | k = 1, 2, ... ,6 \right \}$ together with the probability of each maneuver $P(m_k)$.
The maneuvers are classified by lateral behaviors (including lane-keeping, left and right lane changes) and longitudinal behaviors (including normal driving and braking).
Thereupon, the fused target encoding $\mathcal{T}^+_i$ of target vehicle $v_i\in V_{tar}$ is first fed into a pair of fully connected layers that followed by soft-max layers to get the lateral and longitudinal behavior probability respectively,
and thus their multiplication produces the probability for each maneuver $P(m_k | \mathcal{I}, X)$.
The trajectory under each maneuver class is generated by concatenating the fused target encoding with one-hot vectors of lateral behavior and longitudinal behavior together, followed by passing the resulted feature vector through an LSTM decoder.
Instead of directly generating absolute future locations, our LSTM decoder operates in a residual learning manner that outputs displacement between predicted locations.
The output vector contains the displacement $\delta{y}_i^{t+T} \in \mathbb{R}^2$ between neighboring predicted locations,
the standard deviation vector $ \sigma_i^{t+T} \in \mathbb{R}^2$ and correlation coefficient $\rho_i^{t+T} \in \mathbb{R}$ of predicted location $\hat{y}^{t+T}_{i}$ at the future time step $T \in \left \{ 1, 2, ..., T_{pred} \right \}$.
The predicted location could be accordingly represented by a bivariate Gaussian distribution
\begin{equation}
\hat{y}_i^{t+T} \sim \mathcal{N}(\mu_i^{t+T}, \sigma_i^{t+T}, \rho_i^{t+T}),
\label{eq:normal_distribution}
\end{equation}
where the mean vector is given by summing up all displacements along the future time steps $T$ with the location at the last time step $t$ of history trajectory
\begin{equation}
\mu_i^{t+T} = x_i^t+\sum_{\tau=1}^{T}{\delta{y}_i^{t+\tau}}.
\label{eq:gaussian_mean}
\end{equation}
For brevity, the Gaussian parameters for all future time steps of target $v_i$ is written as $\Theta_i$. Finally, the posterior probability of all target vehicles' future trajectories could be estimated from
\begin{equation}
P(\mathbf{Y}|\mathbf{X}, \mathcal{I}) =
\prod_{v_i\in V_{tar}} \sum_{k=1}^{|M|}
P_{\Theta_i}(Y_i|m_k, \mathbf{X}, \mathcal{I})
P(m_k|\mathbf{X}, \mathcal{I}).
\label{eq:joint_distribution}
\end{equation}

\subsection{Implementation Details}
\label{subsec:implementation}
Our model is trained by minimizing the negative log likelihood of future trajectories under the true maneuver class $m_{true}$ of all the target vehicles
\begin{equation}
- \sum_{v_i\in V_{tar}} \log \left (
P_{\Theta_i}(Y_i|m_{true}, \mathbf{X}, \mathcal{I}) P(m_{true}|\mathbf{X}, \mathcal{I})
\right).
\label{eq:nll}
\end{equation}


Each data instance contains a vehicle specified as the ego.
The predicted targets are the vehicles located within the ego-vehicle-centric area $\mathcal{A}_{tar}$ with the size of $60.96 \times 10.67$ meters ($200 \times 35$ feet), discretized as $25\times5$ spatial grid.
The target-centric area $\mathcal{A}_{nbr}$ of each predicted vehicle is defined the same as $\mathcal{A}_{tar}$.

For the planning input $\mathcal{I}$ of the ego vehicle, its actual trajectory within the prediction horizon is directly used in training.
While in evaluation and testing, $\mathcal{I}$ is fitted from its downsampled actual trajectory.
It is handled in this way because we intend to restrict the prediction from accessing the complete information of planning trajectory, instead only a limited number of waypoints could be accessed. Furthermore, the ground-truth trajectories result from many planning cycles, while in practice, prediction can only be based on the current planning cycle.
So the planning input is represented by a fitted quintic spline, which is a typically used representation for vehicle trajectory.
This feature makes our planning-informed method easy to deploy in a real autonomous system. Although the fitted planning input cannot perfectly fit the actual future trajectory, it could be examined if our method can generalize well in practical use.


%% file: 4_experiments.tex

\section{Experiments}

In this section, we evaluate our method on two publicly available vehicle trajectory datasets, NGSIM~\cite{ngsimDataset} and HighD~\cite{highDdataset}.
Firstly, we compare the performance of our method against the existing state-of-the-art works quantitatively using the metrics of root mean squared error (RMSE) and negative log-likelihood (NLL).
Next, as our method could anticipate different future configurations by performing different plans under the same historical situation, we evaluate PiP from more simulated future situations. Regarding the rationality and variety in generating feasible vehicle trajectories, we employ a model-based vehicle planner MPDM~\cite{cunningham2015mpdm} to generate diverse vehicle trajectories with different lateral and longitudinal behaviors.
In \secref{subsec:user_study}, a user study is conducted by comparing our generated results with the real situations to verify the rationalization of predicted outcomes,
and more results are provided in \secref{subsec:qual_evaluation} for qualitative analysis.


\subsection{Datasets}
\label{subsec:datasets}
We split all the trajectories contained in NGSIM and HighD separately, in which 70\% are used for training with 20\% and 10\% for testing and evaluation.
Each vehicle's trajectory is split into 8s segments composed of 3s of past and 5s of future positions at 5Hz.
The 5s future of ego vehicle used as planning input is further downsampled to 1Hz in testing and evaluation.
The objective is to predict all surrounding target vehicles' future trajectories over 5s prediction horizon.

\textbf{NGSIM:} NGSIM~\cite{ngsimDataset} is a real-world highway dataset which is commonly used in the trajectory prediction task. All vehicle trajectories over a 45-minute time span are captured at 10Hz, with each 15-minute segment under mild, moderate, and congested traffic conditions, respectively.

\textbf{HighD:} HighD~\cite{highDdataset} is a vehicle trajectories dataset released in 2018.
The data is recorded from six different locations on Germany highways from the aerial perspective using a drone. It is composed of 60 recordings over areas of 400 $\sim$ 420 meters span, with more than $110,000$ vehicles are contained.


\subsection{Baseline Methods}
\label{subsec:baseline}
We compare PiP with the following listed deterministic models and stochastic models, and also ablate the planning coupled module and target fusion module in PiP-noPlan and PiP-noFusion respectively, to study their effectiveness in improving predictive accuracy.

\textbf{S-LSTM:}
Social LSTM~\cite{alahi2016slstm} uses a fully connected layer for social pooling and produces a uni-modal distribution of future locations.

\textbf{CS-LSTM:}
Convolutional Social LSTM~\cite{deo2018cslstm} uses convolutional layers with social pooling and outputs a maneuver-based multi-modal prediction.

\textbf{S-GAN:}
Social GAN~\cite{gupta2018sgan} trains GAN based framework using the adversarial loss to generate diverse trajectories for multi-agent in a spatial-centric manner.

\textbf{MATF:}
MATF-GAN~\cite{zhao2019matf} models spatial interaction of agents and scene context by convolutional fusion and uses GAN to produce stochastic predictions.

\subsection{Quantitative Evaluation}\label{quan_evaluation}

Among all the above methods, S-GAN and MATF are stochastic models.\footnote{No NLL results of S-GAN and MATF, as they sample trajectories without generating probability.
No RMSE result of MATF on HighD dataset is reported in \cite{zhao2019matf}.}
We report their RMSE by the best result among 3 samples (i.e., minRMSE).
The others are all deterministic models that generate Gaussian distributions for all predicted locations along the trajectory, in which the means of Gaussian parameters are used as the predicted locations when calculating the RMSE for each time step $t$ within the 5s prediction horizon:
$RMSE(t) = \sqrt{
\frac{1}{\left | V_{tar} \right |}
\sum_{v_i \in V_{tar}}
\left \| y_i - \hat{y}_i \right \|^2
}$.
For multi-modal distribution output by CS-LSTM, PiP and its variants, RMSE is evaluated using the predicted trajectory with the maximal maneuver probability $P(m_k)$.
While RMSE is a concrete metric to measure prediction accuracy, it is limited to some extent since it tends to average all the prediction results and may fail to reflect the accuracy for distinct maneuvers. To overcome its limitation in evaluating multi-modal prediction, we adopt the same way from prior work~\cite{deo2018cslstm} that additionally reports the negative log-likelihood (NLL) of the true trajectories under the predictive results represented by either uni-modal or multi-modal distributions.

\begin{table}[t]
\caption{Quantitative results on NGSIM and HighD datasets are reported by RMSE and NLL metrics over 5s prediction horizon. The best results are marked by bold numbers. Note that for the stochastic methods (S-GAN and MATF), the minimal error from sampling three times reports their RMSE
}
\scriptsize
\begin{center}
\begin{tabular}{@{}cccccccccc@{}}
\toprule
Metric & Dataset & Time
& S-LSTM\cite{alahi2016slstm} & CS-LSTM\cite{deo2018cslstm} & S-GAN\cite{gupta2018sgan} & MATF\cite{zhao2019matf}
& PiP-noPlan & PiP-noFusion & PiP \\
\midrule
\multirow{10}{*}{\rotatebox[origin=c]{90}{RMSE (m)}}
& \multirow{5}{*}{NGSIM}
  & 1s      & 0.60  & 0.58  & 0.57  & 0.66 &\textbf{0.55}   &\textbf{0.55} & \textbf{0.55}  \\
& & 2s      & 1.28  & 1.26  & 1.32  & 1.34 &1.20            & 1.19         & \textbf{1.18}  \\
& & 3s      & 2.09  & 2.07  & 2.22  & 2.08 &2.00            & 1.95         & \textbf{1.94}  \\
& & 4s      & 3.10  & 3.09  & 3.26  & 2.97 &3.01            & 2.90         & \textbf{2.88}  \\
& & 5s      & 4.37  & 4.37  & 4.40  & 4.13 &4.27            & 4.07         & \textbf{4.04}  \\ \cmidrule{2-10}
& \multirow{5}{*}{HighD}
  & 1s      & 0.19  & 0.19  & 0.30  & -    &0.18  &\textbf{0.17}   & \textbf{0.17}  \\
& & 2s      & 0.57  & 0.57  & 0.78  & -    &0.53  & 0.53           & \textbf{0.52} \\
& & 3s      & 1.18  & 1.16  & 1.46  & -    &1.09  &\textbf{1.05}   & \textbf{1.05} \\
& & 4s      & 2.00  & 1.96  & 2.34  & -    &1.86  &\textbf{1.76}   & \textbf{1.76} \\
& & 5s      & 3.02  & 2.96  & 3.41  & -    &2.81  &\textbf{2.63}   & \textbf{2.63} \\
\toprule
Metric & Dataset & Time
& S-LSTM & CS-LSTM & S-GAN & MATF & PiP-noPlan & PiP-noFusion & PiP \\
\midrule
\multirow{11}{*}{\rotatebox[origin=c]{90}{NLL (nats)}}
& \multirow{5}{*}{NGSIM}
  & 1s      & 2.38  & 1.91  & -  & -  &\textbf{1.68} & 1.71          & 1.72  \\
& & 2s      & 3.86  & 3.44  & -  & -  &\textbf{3.29} & \textbf{3.29} & 3.30  \\
& & 3s      & 4.69  & 4.31  & -  & -  &4.20          & \textbf{4.17} & \textbf{4.17}  \\
& & 4s      & 5.33  & 4.94  & -  & -  &4.87          & 4.81          & \textbf{4.80}  \\
& & 5s      & 5.89  & 5.48  & -  & -  &5.42          & 5.33          & \textbf{5.32}  \\ \cmidrule{2-10}
& \multirow{5}{*}{HighD}
  & 1s      & 0.42  & 0.37  & -  & -  &0.20  & 0.20         & \textbf{0.14}  \\
& & 2s      & 2.58  & 2.43  & -  & -  &2.28  & 2.28         & \textbf{2.24}  \\
& & 3s      & 3.93  & 3.65  & -  & -  &3.53  & 3.53         & \textbf{3.48}  \\
& & 4s      & 4.87  & 4.51  & -  & -  &4.39  & 4.37         & \textbf{4.33}  \\
& & 5s      & 5.57  & 5.17  & -  & -  &5.05  & 5.01         & \textbf{4.99}  \\
\bottomrule
\end{tabular}
\end{center}
\label{table:quan_exp}
\end{table}

The results of quantitative results are reported in~\tableref{table:quan_exp}.
Our method significantly outperforms the deterministic models (S-LSTM and CS-LSTM) in both RMSE and NLL metrics on both datasets.
Although sampling more trajectories and choosing the minimal error among all samples would undoubtedly lead to a lower RMSE for stochastic models (S-GAN and MATF), our deterministic model still achieves lower RMSE than stochastic models for sampling three times. The reason for not setting a larger sampling number for the stochastic models is that sampling too many times for prediction may not work well with planning and decision making since the probability of each sample is actually unknown.

The consistent improvements on NLL and RMSE metrics confirm that, by introducing the planning of ego vehicle into the prediction model and capturing the correlations between predictive targets, PiP gets better performance in predictive accuracy.
Additionally, the result of ablation studies shows that PiP-noPlan performs worse than PiP-noFusion in most of the cases, which indicates that the inclusion of the planning trajectory is more effective in improving the accuracy of multi-agent forecasting.


\subsection{User Study}
\label{subsec:user_study}
\begin{figure}[t]
\begin{center}
\includegraphics[width=1.0 \linewidth]{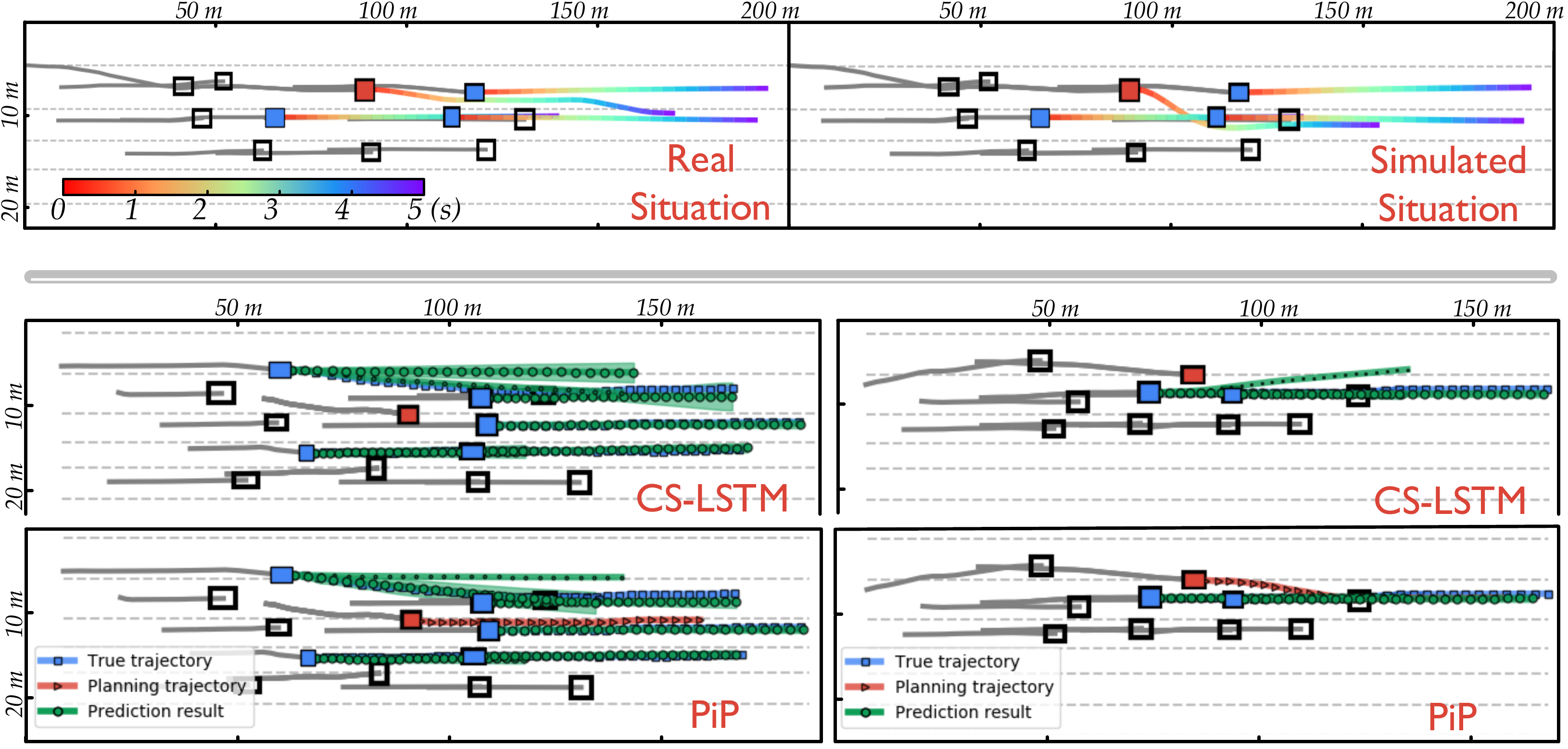}
\end{center}
\caption{
Upper:
user study example of comparing the real and simulated situations.
Each comparison is visualized as video pair for users to choose the situation that violates their intuition.
Lower:
two example cases predicted by CS-LSTM and PiP.
The ground truth (blue), planning (red) and predicted trajectories (green) are visualized by sets of locations with 0.2s time step.
As both methods output maneuver-based multi-modal distributions, only those trajectories with maneuver probability larger than $10\%$ are shown for each target.
The green circle denotes the mean value of distribution on each time step, and its radius is proportional to the maneuver probability of the corresponding trajectory. The green shadow area represents the variance of the distribution.}
\label{fig:compare}
\end{figure}

 To investigate if our prediction model generalizes to various future plans (different maneuver classes and aggressiveness) under different traffic configurations, we have also simulated diverse future scenarios by performing different planned trajectories for the ego vehicle. 
 Accordingly, we conduct the user study that compares real and simulated traffic situations, as shown in the upper part of~\figref{fig:compare}.
Each pair of videos are derived from a segment of 8s traffic sequence recorded in the datasets. One video displays the complete recording of the real tracking data, while the other video shares the same 3s history sequence, and contains a different sequence in the last 5s which is composed by the predicted trajectories of targets (blue) under a different plan performed by ego vehicle (red). The other agents (no color) outside the predictive range are hidden in the last 5s. Note that the same coloring scheme is used in the following experiments.


We display $20$ pairs of videos with randomized order and ask participants to select the one in which the target vesicles' behavior looks unreasonable or against common sense.
Totally $25$ people participated in the user study, and our simulated results were selected as the unreasonable one with a rate of $52.2\% (261/500)$, a bit higher than $50\%$.
One reason is that the ego vehicle's planned trajectory in the simulated results is generated offline, but its real trajectory recorded in the datasets is resulted from replanning adaptively from time to time.
Then it could be a clue for users to select the actual situation as the better one.

Nevertheless, our model still achieves a $47.8\%$ rate of being selected as reasonable. It could also be noted in the upper part of~\figref{fig:compare}, we generate an agile lane merging trajectory for the ego car, and the predicted outcome shows that the following vehicle reacts with deceleration while the leading vehicles maintain speed. Both of the forecastings make sense in real traffic, which indicates that our proposed method could be generalized to different plans.


\subsection{Qualitative Analysis}
\label{subsec:qual_evaluation}
In the following, we further investigate how the prediction is improved as well as explore how PiP enables the planning-prediction-coupled pipeline.

\textbf{Baseline Comparison:} Since our method employs the same maneuver-based decoding as in CS-LSTM~\cite{deo2018cslstm}, the predictive distribution under the same traffic scenes is compared in the lower part of~\figref{fig:compare}.
In the left example, we notice that CS-LSTM outputs similar maneuver probability of keeping the lane and turning right for the left-rear target. At the same time, our method is more confident to target's actual maneuver of turning right.
It is because that ego vehicle is planned to go straight under certain velocity, thereby leaving enough space for the target to merge to its right lane.
By the same token, our method precisely predicts the right-rear target will keep lane but not turn left in the right example. At that moment, the ego vehicle intends to merge to the right lane gradually in a moderate manner, which blocks the way for the right-rear target to turn left in the near future.
Both examples demonstrate that the planning-informed approach leads the prediction to be more accurate.

\begin{figure}[t]
\begin{center}
\includegraphics[width=0.70 \linewidth]{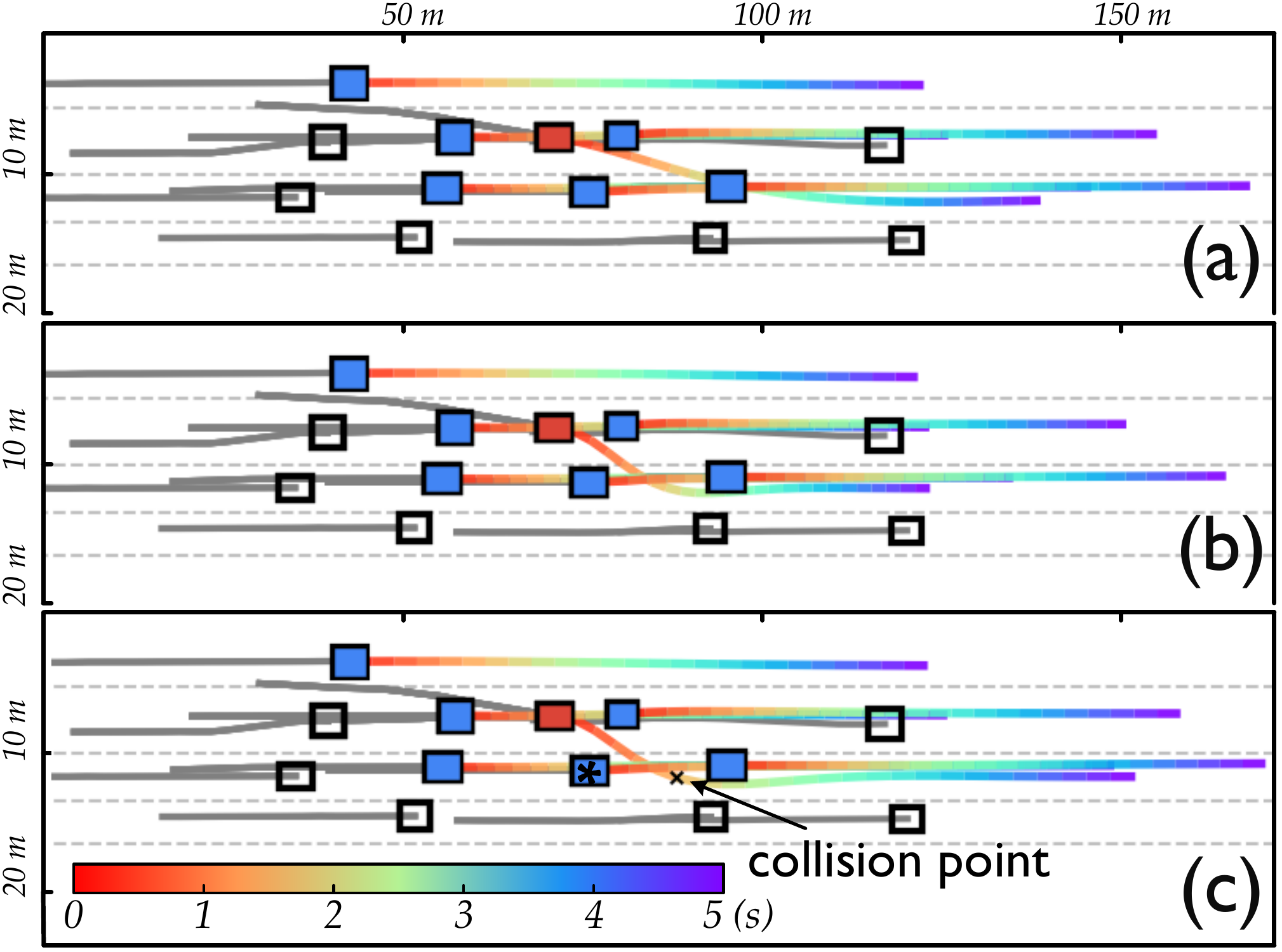}
\end{center}
\caption{Prediction results of performing diverse planned trajectories by ego vehicle: the history trajectories (grey) are from a traffic scene in NGSIM, and the future trajectories are visualized by gradient color varying over time. The target vehicle that collides with ego vehicle is marked with a star symbol, and the collision point is annotated by a cross symbol.}
\label{fig:diff_plan_ngsim}
\end{figure}

\textbf{Active Planning:} With PiP, it is feasible to explore how to plan in different traffic situations actively.
In the following, we illustrate some challenging scenarios with history states acquired from datasets, and PiP produces diverse future states under different plans generated by the ego vehicle.

\figref{fig:diff_plan_ngsim} (a,b) shows prediction results when performing a moderate and aggressive lane changing in dense traffic.
It could be noticed that the aggressive behavior in~\figref{fig:diff_plan_ngsim} (b) is risky as it is very close to the preceding vehicle after merging. Notably, when it merges aggressively a bit faster, as shown in~\figref{fig:diff_plan_ngsim} (c), a collision is forecasted between the controlled vehicle and the target with a star mark.
The ability of forecasting collision further verifies the generalization of our network as no collision occurred in the traffic recordings where the PiP model is trained.

\begin{figure}[t]
\begin{center}
\includegraphics[width=0.95 \linewidth]{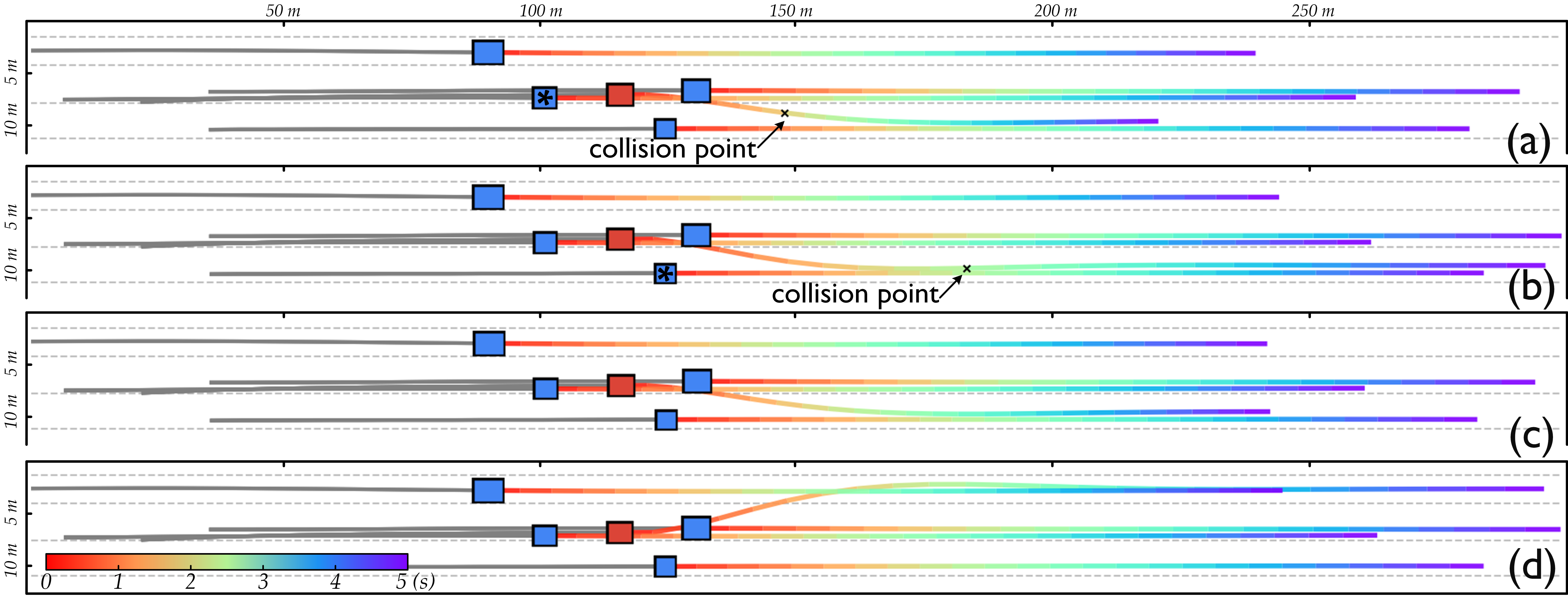}
\end{center}
\caption{Prediction results of performing diverse planned trajectories by ego vehicle: the history trajectories are from a highway scene in HighD. All the annotations are same with~\figref{fig:diff_plan_ngsim}. The predicted future is shown with a collision in (a, b) and safe lane changing in (c,d).}
\label{fig:diff_plan_highd}
\end{figure}

\figref{fig:diff_plan_highd} shows another example from HighD dataset in which the vehicles go much faster than that in NGSIM dataset.
In this case, turning right is challenging. In~\figref{fig:diff_plan_highd} (a) the ego vehicle is planned to turn right and follow the right-front target. A prompt deceleration may cause the rear vehicle to fail to respond and results in a rear-end collision.
PiP also anticipates in~\figref{fig:diff_plan_highd} (b) that a collision will occur if the ego vehicle plans to turn right and overtakes the right-font target.
Nevertheless, it is still possible to find a proper way of merging to the right lane, as shown in~\figref{fig:diff_plan_highd} (c). Additionally, we also show a result of changing to the left lane in~\figref{fig:diff_plan_highd} (d), which is relatively easier as there exists larger space on the left for lane changing.

%% file: 5_conclusion.tex
\section{Conclusion}

In this work, we present PiP for predicting future trajectories in a planning-informed approach. 
Leveraging on the fact that all traffic agents are tightly coupled throughout the time domain, the future prediction on surrounding agents is informed by incorporating history tracks with future planning of the controllable agent. 
PiP outperforms the state-of-the-art works for multi-agent forecasting on highway datasets. 
Furthermore, PiP enables a novel planning-prediction-coupled pipeline that produces future predictions one-to-one corresponding to candidate trajectories, and we demonstrate that it could act as a highly usable interface for planning in dense or fast-moving traffic.
In the future, we plan to extend our approach to work under imperfect tracking or detection information, which is common in the perception module. 
Further, the future prediction and trajectory generation could be integrated into a motion planner that learns to generate optimal planning under interactive scenarios.
